\documentclass[11pt,a4paper]{article}
\usepackage[hyperref]{acl2019}
\usepackage{times}
\usepackage{latexsym}
\usepackage{graphicx}
\usepackage{multirow}
\usepackage{tabularx}
\usepackage{amsmath}
\usepackage{algorithmic}
\usepackage[ruled]{algorithm2e}
\usepackage{tikz}
\usepackage{makecell}
\usepackage{comment}
\usepackage{amssymb}
\usepackage{arydshln}
\usepackage{enumitem}
\usepackage{cancel}
\usepackage{array,multirow,graphicx}
\usepackage{pgfplots}

\aclfinalcopy

\newcommand\BLEU{\textsc{Bleu}\xspace}

\title{Dynamically Composing Domain-Data Selection with Clean-Data Selection by ``Co-Curricular Learning''
for Neural Machine Translation}
\author{Wei Wang \and Isaac Caswell  \and Ciprian Chelba \\ 
  Google Research \\
  \texttt{\{wangwe,icaswell,ciprianchelba\}@google.com} \\
  }

\date{}

\begin{document}
\maketitle
\begin{abstract}
Noise and domain are important aspects of data quality for neural machine translation. Existing research focus  separately on domain-data selection, clean-data selection, 
or their static combination, leaving the dynamic interaction across them
not explicitly examined.
 This paper introduces a ``co-curricular learning'' method
to compose dynamic domain-data selection with dynamic clean-data selection, for transfer
learning across both capabilities.
We apply an EM-style optimization procedure to further refine the ``co-curriculum''.
Experiment results and analysis with two domains demonstrate the effectiveness of the method
and the properties of data scheduled by the co-curriculum.

\begin{comment}
Noise and domain are important aspects of data quality for neural machine translation.
Research has focused separately on domain-data selection, or clean-data selection, 
or their static combination, leaving the dynamic interaction across them
not explicitly examined. This paper introduces a ``co-curricular learning'' method to
compose dynamic domain-data selection with dynamic clean-data selection, to transfer both capabilities into a final co-curriculum that is better than either constituent one or
their static combination, particularly on noisy parallel data.
We introduce an optimization procedure to bootstrap the composed curriculum,
and gain additional improvements, especially with small models. 
\end{comment}

%an effect usually seen in model distillation, without a teacher in our case.

% introduces a ``co-curricular learning'' method to compose dynamic domain-data selection
%with dynamic clean-data selection,
%to transfer both capabilities into a final ``co-curriculum''. 
%Our method takes as input a small source monolingual corpus that is in-domain, and a small, trusted but
%out-of-domain parallel corpus. It then schedules training examples from a large, background dataset to train
%an NMT model. 
%We  apply an EM-style optimization procedure
%to further refine the  co-curriculum.
%Results show that NMT models trained by our method translate the in-domain data  better than models trained on either of the original data selection do, in particular for domain-data selection on noisy parallel corpora.

\end{abstract}

\section{Introduction}
Significant advancement has been witnessed in neural machine
 translation (NMT), thanks to  better modeling and data.
 As a result, NMT has found  successful  use cases in, for example, domain translation and helping other NLP applications, e.g., \cite{ActiveQA,NIPS2017_7209}.  As these tasks start to scale to more
 domains, a  challenge starts to surface: Given a source monolingual corpus,
 how to use it to improve an NMT model to translate same-domain sentences
   well? Data selection  plays an important role in this context.

\begin{figure}[t]
\begin{center}
\includegraphics[scale=0.5]{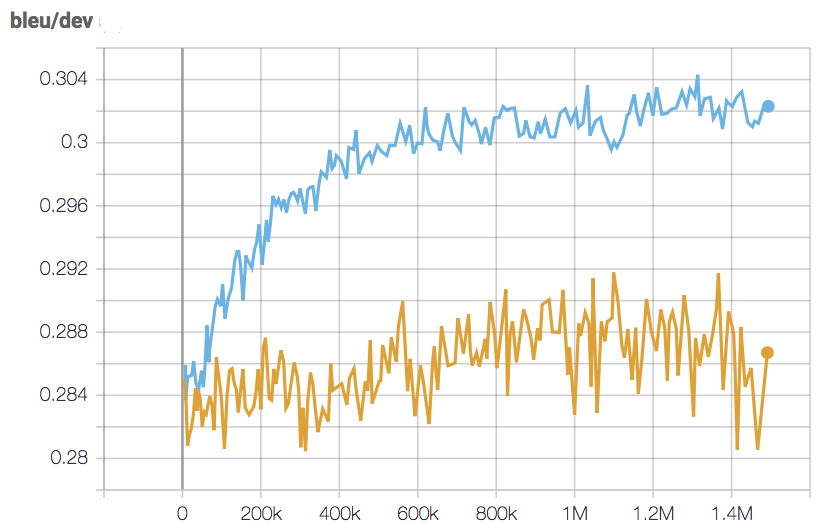}
\end{center}
  \caption{
  \small{
\BLEU  curves over NMT training steps:  domain-data selection on Paracrawl  English$\to$French data
 (lower curve) vs.   clean-data selection 
 on the same   data (upper curve).  Setup available in  the experiment section.}
% Query domain is WMT news.
\label{fig1}
}
\end{figure}

In machine translation, data selection has been a fundamental 
 research topic. One idea \cite{dynamiccds,Axelrod2011}
 for this problem is to use language models to select parallel data out of a background
 parallel corpus, seeded by the  source monolingual sentences.
 This approach, however, performs poorly on noisy data, such as large-scale, web-crawled datasets,
 because data noise  hurts NMT performance \cite{nmtnoise}.
 The lower learning curve in Figure~\ref{fig1} shows
 the effect of  noise on domain-data selection.
  %%%%%%%%in  both translation accuracy and training convergence speed

NMT  community has realized the harm of data noise to translation quality,
leading to efforts in data denoising \cite{koehn-EtAl:2018:WMT}, as has  been popular in computer vision \cite{trusted_data_denoise}.  The upper curve in Figure~\ref{fig1} shows the effect of  clean-data selection on the same noisy data. These denoising methods, however, cannot be directly used for the  problem in question as they require  trusted parallel data as input.
 
\begin{comment}
Denoising has been popular in  computer vision, too \cite{trusted_data_denoise}. A commonly used technique there
is to add a confusion matrix to corrupt the softmax distribution to a noisy distribution that matches the noisy data.
As a result, this layer absorbs the noise and the inner softmax is pushed to a clean distribution.

Curriculum learning (CL) \cite{curri_learn} has gained attention in NMT, for example,
to improve static data selection by the dynamic
interaction between data and network \cite{dynamiccds,denoise_nmt}. Most existing data-selection CL
work are concerned with {\em in-task} dynamic example scheduling,
for either domain or denoising. There has been static combination, but cross-task curriculum interaction
has not been investigated.
\end{comment}

We introduce a method to dynamically combine clean-data selection and
domain-data selection. We treat them as  independent curricula, and  compose them
into a  ``co-curriculum''. 
%We further apply an EM-style
%optimization procedure to refine the co-curriculum.
We summarize our contributions as:

\begin{enumerate}
\item ``Co-curricular learning'', for transfer learning across data quality.
%a simple method to dynamically compose domain-data selection
%with clean-data selection. 
It extends the single curriculum learning work in NMT and
makes the existing domain-data selection method  work better with noisy 
data.

\item A curriculum optimization procedure to refine the co-curriculum. While gaining some 
improvement with deep models, it surprisingly improves shallow model by 8-10 BLEU points -- 
We find that bootstrapping seems to ``regularize'' the  curriculum and make it easier for a small model to learn on. 

\item We wish our work contributed towards better understanding of data, such as noise, domain,  or ``easy to learn'', and its interaction with NMT network. 

\end{enumerate}

\section{Related Work}
\subsection{Measuring Domain and Noise in Data}
Data selection for MT usually uses a scoring function to rank sentence pairs.
Cross entropy difference \cite{Moore2010} between two language models is usually used for
selecting domain sentences, e.g.,  \cite{dynamiccds,Axelrod2011}. 
 For a source sentence $x$ of length $|x|$, with  a general-domain language model (LM), parameterized as $\widetilde{\vartheta}$,
 and an in-domain LM, $\widehat{\vartheta}$, the domain-relevance of $x$ is calculated as:\footnote{
 We can use both source and target LMs, but we study the problem where  only a source in-domain corpus is available..
 }
\begin{eqnarray}
\varphi\left(x; \widetilde{\vartheta}, \widehat{\vartheta}\right) = 
   \frac{\log P\left(x;\widehat{\vartheta}\right) - \log P\left(x; \widetilde{\vartheta}\right)}{|x|} \label{varphi}
\end{eqnarray}
%\citet{dynamiccds} implement LMs using neural networks, in which case,  $\widehat{\vartheta}$ and  $\widetilde{\vartheta}$ are the network weights.
 Alternative   measures \cite{D17-1155,chen2016a,chen2016b} also show  effectiveness.
 With Eq.~\ref{varphi} to select data, the data distribution (domain quality)  in the in-domain
 monolingual data used to train $P(x; \widehat{\vartheta})$ is transferred into the selected data through the scoring.

Data selection has also been used for data denoising \cite{junczysdowmunt:2018:WMT2,denoise_nmt}, 
by using NMT models and trusted data to measure the noise level in a sentence pair.
One such a scoring function uses a baseline  NMT, $\widetilde{\theta}$, trained on  noisy data
and a cleaner NMT, $\widehat{\theta}$, obtained by fine-tuning $\widetilde{\theta}$
 on a small trusted parallel dataset,
and measures quality in  a sentence pair $(x, y)$:
 \begin{eqnarray}
\phi\left(x,y; \widetilde{\theta}, \widehat{\theta}\right) \hspace{-0.07in} = \hspace{-0.05in}
 \frac{\log P\left(y|x;\widehat{\theta}\right) \hspace{-0.02in} - \hspace{-0.02in} \log P\left(y|x; \widetilde{\theta}\right)}{|y|} \hspace{-0.05in} \label{phi}
\end{eqnarray}
Using NMT models for selection can also lead to faster convergence \cite{wang-etal-2018-dynamic}.
With Eq.~\ref{phi}, the distribution (data quality) in the trusted parallel data is transferred into the selected data. These scoring functions  usually use smaller networks.

%[Data filtering task has used both LM and NMT to filter data, without study the interaction between these two.]

%[Data quality (distribution) transfer from the supervision data into the selected data]

%. In this task,  the amount of noise in a sentence pair is measured with a noisy NMT model and 
%a denoised NMT model trained on a small amount of trusted parallel data. 

 %Online data selection has been recently used for the {\em task of denoising} NMT training. \cite{denoise_nmt} measure noise in a sentence pair with a noisy NMT model and a denoised NMT model trained on a small amount of trusted parallel data. Data is then ordered from noiser to cleaner to achieve the denoising purpose. The noise in an example is measured in a similar way to how domain relevance is measured. 
 
 \subsection{Curriculum Learning for NMT}
 %\cccomment{I still find this description here too far from what you describe in Sections 3 and 4. Is $P(y|x)$ in Eq. 3 the training data distribution? If so, it should be joint; usually people use $f(x,y)$ for that since it is a relative frequency. The weighting function $W_t(x,y)$ needs to be described clearly as a 0/1 windowing function operating on the ranking produced by some scoring function $\phi(x,y)$. The actual scoring function used, the windowing function, the windowing schedule with t should all be described either here or in Section 3 if this is a novel/non-standard approach. A sentence explaining that in fact this amounts to simply sampling at random from the top of the ranking with more and more restrictive thresholds as training steps t progress would also clarify things.}
Curriculum learning (CL) \cite{curri_learn} has been used to further improve traditional static selection.
In CL, a curriculum, $\mathcal{C}$,
is a sequence of training criteria over training steps.
 A training criterion, $Q_t(y|x)$, at step $t$ is associated with a set of weights,
 $W_t(x,y)$, over training examples $(x,y)$ in a dataset $D$, where
 $y$ is the translation for $x$. $Q_t(y|x)$ is a re-weighting of the training distribution $P(y|x)$:
\begin{equation} 
\begin{split}
Q_t\left(y|x\right) \propto W_t\left(x,y\right)P\left(y|x\right), 
  \hspace{-0.1in} \hspace{0.1in} \forall (x,y) \in D \label{form1}
\end{split}
\end{equation}
Hence, for a training with $T$ maximum steps, $\mathcal{C}$ is a sequence:
\begin{eqnarray} 
  \mathcal{C} = \left\langle Q_1, ..., Q_t, ..., Q_T \right\rangle  \label{form2}
\end{eqnarray}
 At  $t$, an online learner samples data from $Q_t$ to train on, resulting
 in a task (or model), $m_t$. Therefore,  $\mathcal{C}$ corresponds to a sequence of tasks,
 $\mathcal{M} = \left\langle m_1, ..., m_t..., m_f \right\rangle$,
 %Given $D$,  ${W_t(x, y), $0 < t \le T}$ fully characterizes $\mathcal{C}$. 
%Curriculum learning is a form of transfer learning, which transfers knowledge
 %through a curriculum, in order to benefit a final task. 
 %Therefore,                                  
%\begin{eqnarray}
%\mathcal{M} = \langle m_1, ..., m_t..., m_f \rangle.
%\end{eqnarray}
%with $m_t$ being learned from $Q_t$.  
where $m_f$ is the final task of interest. 
% which is referred to when we talk about the task of the curriculum.
 Intermediate tasks,  $m_t$, are sorted in increasing
 relevance to $m_f$  as a series of ``stepping stones''  to $m_f$, 
 making curriculum learning  a form of transfer learning that transfers knowledge
 through $\mathcal{M}$ to benefit $m_f$. A performance metric
 $\mathcal{P}(\mathcal{C}, m_f)$ is used to evaluate $m_f$.
 %after training on $\mathcal{C}$.
  
 %A task-relevance measure is used to order data, thus sequencing the tasks.
 %Its accuracy is crucial for a curriculum to optimize the right final task of interest. 
 %Historically, curriculum learning has been studied with a single final task.
 %A popular data order  starts from easy examples in a data subset, progressively widened to
 % difficult examples in the full data, usually for faster convergence. 
  %\vspace{0.05in}

  There has already been rich research in CL for NMT. Fine-tuning a baseline on in-domain parallel data
 is a good strategy \cite{W18-6313,model_stacking, DBLP:journals/corr/FreitagA16}. 
\citet{dynamiccds} introduce a domain curriculum.
\citet{denoise_nmt} define noise level and introduce a  denoising curriculum.
\citet{R17-1050} use linguistically-motivated features to classify examples into bins for scheduling. 
\citet{curriculum_opt_for_nmt} use reinforcement learning to learn a denoising curriculum based on noise level of examples.
% A similar idea has also been used to improve noise robustness for speech recognition \cite{DBLP:journals/corr/BraunNL16}.
%  All these show significant accuracy improvements.
  % by using curriculum learning.
  \citet{DBLP:journals/corr/abs-1811-00739} explore CL in general for NMT and observe faster training convergence. 
\citet{curriculum_domain} use CL to adapt generic NMT models to a specific domain.
\citet{DBLP:journals/corr/abs-1903-09848} propose a CL framework to simplify and speed up training and achieve better results; a nice study in sampling schedules was carried out.
%Curriculum learning has also been employed to speed up  training.  
%\citet{DBLP:journals/corr/abs-1811-00739} show faster NMT training convergence  and point out that
%the learning process is  sensitive to the task relevance measure, among other factors. 
%This confirms that a good data-relevance measure  is important to curriculum learning. 
%In most these works, ``planning-based'' curricula are used where the curricula are manually designed,
% in a self-paced manner \cite{NIPS2010_3923}. 
 %\vspace{0.05in}

CL therefore is a natural formulation for dynamic online data selection. 
Our work is built on two types of dynamic data selection:
Dynamic domain-data selection and dynamic clean-data selection.
The former uses the neural  LM (NLM)-based scoring function (Eq.~\ref{varphi}), which we call {\bf domain curriculum}, denoted by $\mathcal{C}_\mathrm{domain}$.
The later uses the NMT-based scoring function (Eq.~\ref{phi}), which we call {\bf denoising curriculum}, denoted by $\mathcal{C}_\mathrm{denoise}$.
Ideally, we would
have in-domain, trusted parallel data to design a {\bf true curriculum},  $\mathcal{C}_{\mathrm{true}}$, as an assessment oracle:
 with trusted in-domain parallel data,   $\mathcal{C}_{\mathrm{denoise}}$ 
is expected to simultaneously perform domain-data selection and clean-data selection, becoming $\mathcal{C}_{\mathrm{true}}$.

Mini-batch sampling is  important for CL. Several alternatives have been introduced 
 to evolve  the
training criteria $Q_t$ over time \cite{DBLP:journals/corr/abs-1811-00739,denoise_nmt,dynamiccds,R17-1050,DBLP:journals/corr/abs-1903-09848}. 
In these curricula, tasks in $\mathcal{M}$ are sequenced in order of increasing
 relevance. Earlier tasks are exposed to a diversity of examples
%to encourage exploration. $Q_t$
and later tasks progressively concentrate on  data subsets more relevant to the final task.

 %Each piece of work concerns  a single final task.

%Building on this exiting NMT research, our work is concerned with composing two partial curricula,
%a domain curriculum and a denoising curriculum, to transfer their respective capabilities into another  that
%does both. We refer to this as ``co-curricular learning''. In
%the learning process, it aggregates two (final) tasks into one.
%It has its inspiration in composing Q-network policies in reinforcement
% learning \cite{DBLP:conf/icra/HaarnojaPZDAL18}. 

\subsection{More Related Work}
 \citet{junczysdowmunt:2018:WMT2} introduces a practical and effective method to combine (static) features
 for data filtering.   \citet{LM_M1} combine an n-gram LM and IBM translation Model 1 \cite{Brown:1993:MSM:972470.972474} for domain data filtering.
 We compose different types of dynamic online selection rather than combining static features.

Back translation (BT), e.g., \cite{sennrich-haddow-birch:2016:P16-11}, is another important approach to using monolingual data for NMT.
% particularly for domain adaptation. 
Here we use monolingual data to seed data selection, rather than generating parallel data directly from it. Furthermore, we study the use of source-language monolingual data, in which case BT cannot be applied directly.

\section{Problem Setting}  \label{problem}
$\widetilde{D_{XY}}$ is a background parallel dataset   between languages $X$ and $Y$. It may be crawled from the web: large (hundreds of millions of pairs),
  diverse
 and  noisy.
 %It can contain hundreds
 %of millions of sentence pairs or more. 
 %It is the  dataset we select data from.
 
 $D_{X}^{\text{ID}}$ is an in-domain  monolingual corpus in source language
 $X$. It contains thousands to millions of
 sentences and specifies the testing domain.
With $D_X^{\text{ID}}$, we can train $\varphi$ (Eq.~\ref{varphi})  to sort data by domain relevance into a domain
 curriculum. $D_{X}^{\text{ID}}$ can be small because we can use it to fine-tune $\widetilde{\vartheta}$
 into $\widehat{\vartheta}$.

$\widehat{D_{XY}^{\text{OD}}}$ is a small, trusted, out-of-domain (OD) parallel dataset.
  It contains several thousands of
 pairs or fewer.  With $\widehat{D_{XY}^{\text{OD}}}$,  we can train
 the $\phi$ (Eq.~\ref{phi}) to sort data by noise level  into  a denoising
 curriculum.
 %or form a noise correction layer.
 
 The setup, however, assumes that the in-domain, trusted parallel data, $\widehat{D_{XY}^{\text{ID}}}$, does not exist -- Our goal
 is to use an easily available monolingual corpus and recycle existing trusted parallel data
to reduce the cost of curating in-domain parallel data.

We are interested in a composed curriculum, $\mathcal{C}_\mathrm{co}$, to improve  either original  curriculum:
 \begin{eqnarray}
      \mathcal{P}\left(\mathcal{C}_{\text{co}}, m_f\right) &>& \mathcal{P}\left(\mathcal{C}_{\text{denoise}}, m_f\right) \label{p1}           \\
         \mathcal{P}\left(\mathcal{C}_{\text{co}}, m_f\right) &>& \mathcal{P}\left(\mathcal{C}_{\text{domain}}, m_f\right) \label{p2}
 \end{eqnarray} 
We hope $\mathcal{P}(C_{\text{co}}, m_f) \approx \mathcal{P}(\mathcal{C}_\mathrm{true}, m_f)$ 
as if a small in-domain, trusted parallel dataset were available.

%Our goal is to transfer the domain-discerning capability learn from $D_{X}^{\text{ID}}$ and the
%denoising capability learned from $D_{XY}^{\text{OD}}$ into a new task that improves either of the
%capability.
 % $\mathcal{C}_{\text{denoise}}$.

  %  $\mathcal{C}_{\text{query}}$, to train a domain NMT.

\begin{comment}
In principle, the most desirable setup is to curate some in-domain,
 trusted parallel dataset, $D_{XY}^{\text{ID}}$, in place of $D_X^{\text{OD}}$, in order
 to form a bilingual-domain curriculum, $\mathcal{C}_{\text{true}}$.
 However, the co-curriculum solution we are seeking for has an obvious
 advantage over the bilingual curriculum in both scalability and cost effectiveness,
 thus is more attractive. Of course, in addition to requirements~\ref{p1} and~\ref{p2}, $\mathcal{C}_{\text{co}}$
is also desired to be
\begin{eqnarray}
 \mathcal{P}(\mathcal{C}_{\text{co}}, m_f)  \approx \mathcal{P}(\mathcal{C}_{\text{true}}, m_f). \label{p3}
\end{eqnarray}
\end{comment}

\section{Co-Curricular Learning} \label{our_approach}

%%%%% , in place of $D_X^{\text{OD}}$:
%\begin{eqnarray}
% \mathcal{P}(\mathcal{C}_{\text{co}}, m_f)  \approx \mathcal{P}(\mathcal{C}_{\text{true}}, m_f). \label{p3}
%\end{eqnarray}

\begin{table}[t]
\small
\centering
%  \begin{tabularx}{\textwidth}{@{}@{}l|lX@{}}
\begin{tabularx}{\textwidth}{@{}llX@{}}
    %\hline
%\multirow{6}{*}{3 en-$>$zh sentence pairs} &\\
&\underline{3 en$\rightarrow$zh sentence pairs:} \\
      \multirow{2}{*}{1} &(en) Where is the train station? \\
                           &(zh-gloss) {\small TRAIN STATION IS WHERE?} \\
     \multirow{2}{*}{2} &(en)  I’d like to have two window seats. \\
                          &(zh-gloss) {\small PLEASE BOOK ME TWO WINDOW SEATS.} \\
     \multirow{2}{*}{3} &(en) It usually infects people older than 60. \\
                           & \makecell[lp{2.8in}]{(zh-gloss) \small{ PEOPLE OLDER THAN 60 USUALLY ARE INFECTED BY IT.}} \\
   % \hline
   \\
\multicolumn{2}{l}{     \hspace{1.4in} $W_1$ $\hspace{0.03in}\to\hspace{0.03in}$   $W_2$ $\hspace{0.03in}\to\hspace{0.03in}$  $W_3$ $\hspace{0.03in}\to\hspace{0.03in}$   $W_4$} \\
%\multicolumn{2}{l}{\makecell[l]{Travel domain curriculum \\ \hspace{0.1in} $\varphi(3)<\varphi(2)<\varphi(1)$} } \hspace{-1in}
&\makecell[l{{p{2.8cm}}}]{Travel domain curri. \\ $\varphi(3)<\varphi(2)<\varphi(1)$ }
                $\begin{pmatrix}1/3\\1/3\\1/3\end{pmatrix}$
    	            $\begin{pmatrix}1/3 \\ 1/3 \\ 1/3\end{pmatrix}$ 
                $\begin{pmatrix}1/2 \\ 1/2 \\ \cancel{0.0} \end{pmatrix}$ 
                $\begin{pmatrix}1.0\\\cancel{0.0}\\\cancel{0.0}\end{pmatrix}$  \\
               % $\begin{pmatrix}1.0 \\ \cancel{0.0} \\ \cancel{0.0} \end{pmatrix}$  \\
        %  \hdashline
%\multicolumn{2}{l}{\makecell[l]{Denoising curriculum \\ \hspace{0.1in}  $\phi(2)<\phi(1)<\phi(3)$} }
&\makecell[l{{p{2.8cm}}}]{Denoising curri. \\ $\phi(2)<\phi(1)<\phi(3)$ }
    $\begin{pmatrix}1/3\\1/3\\1/3\end{pmatrix}$ 
    $\begin{pmatrix}1/2\\\cancel{0.0}\\1/2\end{pmatrix}$ 
     $\begin{pmatrix}1/2\\\cancel{0.0}\\1/2\end{pmatrix}$ 
      $\begin{pmatrix}1/2 \\  \cancel{0.0}\\\ 1/2 \end{pmatrix}$ \\
       
  %  $\begin{pmatrix} \cancel{0.0} \\ \cancel{0.0} \\ 1.0 \end{pmatrix}$  \\
  %  \hdashline

%\multicolumn{2}{l}{\makecell[l]{Co-curriculum \\ \hspace{0.1in} (Our goal)}    
&\makecell[l{{p{2.8cm}}}]{Co-curriculum \\  (Our goal)}   
          $ \begin{pmatrix}1/3 \\ 1/3 \\ 1/3\end{pmatrix}$ 
   $ \begin{pmatrix}1/2\\ \cancel{0.0} \\ 1/2 \end{pmatrix}$ 
   $ \begin{pmatrix}1.0 \\ \cancel{0.0} \\ \cancel{0.0} \end{pmatrix}$ 
   $ \begin{pmatrix}1.0 \\ \cancel{0.0} \\ \cancel{0.0} \end{pmatrix}$ \\
   %$ \begin{pmatrix} \cancel{0.0} \\ \cancel{0.0} \\ \cancel{0.0} \end{pmatrix}$ 
     \\
%\hline
  \end{tabularx}
\caption{\small Curriculum and co-curriculum examples generated from a toy dataset. Each is characterized by its re-weighting, $W_t$, 
over four steps, to stochastically order data to benefit a final task. $\varphi$: the domain scoring function (Eq.~\ref{varphi}).
$\phi$: the denoising scoring function (Eq.~\ref{phi}). Strikethrough marks discarded examples. 
%Pace function floor values
%prevent empty $W_t$.
 \label{curriculum}}
%\label{CL}}
\end{table}
 Table~\ref{curriculum} illustrates the idea with a toy dataset of three examples. Source sentences (en) of examples 1 and 2
 are in the travel domain. Example 2 is a  noisy translation.  Example 3 is well-translated but belongs to the medicine domain.
 A travel-domain curriculum follows its data re-weighting, $W_t$, and gradually discards (strikethrough)
 less in-domain examples,  optimizing towards a travel-domain model. The denoising curriculum gradually discards noisy examples  to 
 improve general accuracy, without paying special attention to  travel domain. We want to ``fuse'' these two partial curricula into a
 co-curriculum to train  models progressively on  both in-domain and clean examples. We call this co-curricular learning.

 %Our goal is to automatically determine a curriculum over these three examples that optimizes for both domain and quality.
 
 %We introduce two different types of  co-curricula below.
 \begin{comment}
We use $\mathcal{C}_\mathrm{denoise}$ to denoise $\mathcal{C}_\mathrm{query}$ by
composing them in one NMT training. We refer to training NMT
on the composed co-curriculum  as co-curricular learning. 
The co-curriculum needs to be constructed to both select domain and denoise,
resulting in better performance  than either of the original curricula.
%According to Eq~\ref{form1}, curriculum is fully characterized by the re-weighting, $W_t(x,y)$, we therefore
%define the co-curricula by defining $W_t(x, y)$. 
%We introduce two co-curricula below.
\end{comment}

 %With $D_{XY}^{\text{OD}}$,  we can form a denoising
% curriculum, $\mathcal{C}_{\text{denoise}}$, using Formulae~\ref{ds_as_curri}
% and~\ref{phi}.  $\mathcal{C}_{\text{denoise}}$ sequences tasks
 %by letting them focus on progressively denoised data, in order
 %to yield a final model that translates generally well across different test cases.
 
% With $D_X^{\text{ID}}$, we can form a query-domain
 %curriculum,  $\mathcal{C}_{\text{query}}$, to train a domain NMT,
 %using Formulae~\ref{ds_as_curri} and \ref{varphi}. $\mathcal{C}_{\text{query}}$’s
 %performance may be severely compromised by the data noise in $D_{XY}^{\text{background}}$.
 
 \subsection{Curriculum Mini-Batching} 
 To facilitate the definition of co-curricular learning and following \cite{DBLP:journals/corr/abs-1903-09848,denoise_nmt},
 we define a dynamic data selection  function, $\mathcal{D}_\lambda^{\phi}(t, D)$, 
 to return the top $\lambda(t)$  of  examples in a dataset $D$ sorted by
 a scoring function $\phi$ at a training step $t$. We use $\lambda(t)=0.5^{t/H}$,  ($0 < \lambda \le 1 $), as a {\em
 pace function} to return a selection ratio value that decays
 over time controlled by a  hyper-parameter $H$.\footnote{
 This is inspired by the exponential learning rate schedule.
 In the following notations, we omit $H$ for brevity, but the function name implies it.
 }
 During training, $\mathcal{D}_\lambda^{\phi}(t, D)$ progressively evolves into smaller subdatasets that are more relevant
 to the final task using the scoring function. In practice,  $\mathcal{D}_\lambda^{\phi}(t, D')$ can be applied
 on a small buffer $D'$ of random examples from the much bigger $D$,  for efficient online  training. It may also be
 desirable to set a  floor value on $\lambda(t)$ to avoid potential data selection bias.
 This is how we implement a curriculum in experiments.
  We introduce two different  co-curricula below.

%\subsection{Co-Curriculum by  Addition ($\mathcal{C}_{\text{co}}^{\text{\small{add}}}$)}
\subsection{Mixed Co-Curriculum ($\mathcal{C}_{\text{co}}^{\text{\small{mix}}}$)}
Mixed co-curriculum, $\mathcal{C}_{\text{co}}^{\text{\small{mix}}}$, simply adds up the 
domain scoring function (Eq.~\ref{varphi}) and the denoising function (Eq.~\ref{phi}).
%Co-curriculum $\mathcal{C}_{\text{co}}^{\text{\small{Sum}}}$ define the scoring function,
%$\phi^{\text{\small{Sum}}}(x, y)$, by adding up the partial curricula's scoring functions. 
 For a sentence pair $(x, y)$,
\begin{eqnarray*}
\psi(x, y) = \phi(x, y)+\varphi(x).
\end{eqnarray*}
%where $\phi$ and $\varphi$ are defined in Eq.~\ref{varphi} and Eq.~\ref{phi}, respectively.
%We denote this co-curriculum by $\mathcal{C}_{\text{co}}^{\text{\small{add}}}$.
%Composing scoring functions of curricular this way can find its 
%inspiration from research, e.g,  \cite{DBLP:conf/icra/HaarnojaPZDAL18}, in 
%combining Q-network policies for reinforcement learning.
%With a selection function, $\mathcal{D}_\lambda^\psi(t, \widetilde{D_{XY}})$,  applied on the  background data,
We then can constrain the re-weighting, $W_t(x, y)$, to assign non-zero weights only to examples in $D_{\lambda}^{\psi}(t, \widetilde{D_{XY}})$
at a training step.  We use uniform sampling.
%That is,
 \begin{comment}
 \begin{equation*}
 W_{t}\left( x,\ y\right) =\begin{cases}
\frac{1}{|\mathcal{D}_{\lambda}^{\psi}(t, \widetilde{D_{XY}})|}  &\hspace{-0.1in} \mathrm{if} \ (x,y) \hspace{-0.05in}\in \hspace{-0.05in}\mathcal{D}_\lambda^\psi(t, \widetilde{D_{XY}}) \\
0  & \hspace{-0.1in} \mathrm{otherwise} 
\end{cases} 
\end{equation*}
where \widetilde{D_{XY}} is the background data to select on.
\end{comment}
The co-curriculum is thereby fully instantiated based on Eq.~\ref{form1}~and Eq.~\ref{form2}. 
However, values of  $\phi$ and $\varphi$ may not be on the same scale or even from the same family of distributions. Therefore, despite
its simplicity,  $\mathcal{C}_{\text{co}}^{\text{\small{mix}}}$ may not be able to
enforce either curriculum sufficiently.

\subsection{Cascaded Co-Curriculum ($\mathcal{C}_{\text{co}}^{\text{\small{cascade}}}$)} \label{nested_cocurriculum}
Cascaded co-curriculum, $\mathcal{C}_{\text{co}}^{\text{\small{cascade}}}$, defines two selection functions and nests them.
Let  $\beta\left(t\right)=0.5^{t/F}$ and $\gamma\left(t\right)=0.5^{t/G}$ be two pace functions,  implemented similarly to above
 $\lambda(t)$, with different  hyper-parameters $F$ and  $G$.\footnote{
 We will omit $F, G$ for brevity, but the function names can indicate them.
 } They control the data-discarding paces for clean-data selection and domain-data selection, respectively.
 At step $t$,  $\mathcal{D}_{\beta}^{\phi}\left(t, \widetilde{D_{XY}}\right)$ retains the top $\beta\left(t\right)$ of background data
 $\widetilde{D_{XY}}$, sorted by scoring function $\phi\left(x, y\right)$.
 $\mathcal{D}_{\gamma}^{\varphi}\left(t, \mathcal{D}_{\beta}^{\phi}\left(t, \widetilde{D_{XY}}\right)\right)$ retains
the top $\gamma\left(t\right)$ of  $\mathcal{D}_{\beta}^{\phi}\left(t, \widetilde{D_{XY}}\right)$, re-sorted by scoring function $\varphi\left(x\right)$.
That is,
\begin{equation*}
\left(\mathcal{D}_{\gamma}^{\varphi} \circ \mathcal{D}_{\beta}^{ \phi}\right)\left(t, \widetilde{D_{XY}}\right) 
%\mathcal{D}_{\gamma \circ \beta}^{\varphi \circ \phi}(t, D_{XY}^{\mathrm{bg}}) 
= \mathcal{D}_{\gamma}^{\varphi}\left(t, \mathcal{D}_{\beta}^{\phi}\left(t, \widetilde{D_{XY}}\right)\right) 
\end{equation*}
%where $\widetilde{D_{XY}}$ is the background parallel data to select on.
Then Eq.~\ref{form1} is redefined into Eq.~\ref{form2} with uniform sampling:\footnote{
Function nesting is asymmetrical, but the uniform sampling seems to make the nesting irrelevant to the nesting order. In experiments, we did
not notice empirical differences between nesting one way or the other.
}
 %$D_{\gamma \circ \beta}$:
\begin{eqnarray}
W_t\left(x, y\right) = 
 \begin{cases}
%\frac{1}{|\mathcal{D}_{\gamma \circ \beta}^{\varphi \circ \phi}|}  &  \mathrm{if} \ (x,y) \in  \mathcal{D}_{\gamma \circ \beta}^{\varphi \circ \phi} \\
\frac{1}{|\mathcal{D}_{\gamma}^\varphi \circ \mathcal{D}_{\beta}^{ \phi}|}  &  \mathrm{if} \ \left(x,y\right) \in \mathcal{D}_{\gamma}^\varphi \circ \mathcal{D}_{\beta}^{ \phi} \\
0  & \mathrm{otherwise}
\end{cases} \label{uniform}
\end{eqnarray}
 Compared to $\mathcal{C}^\mathrm{mix}_\mathrm{co}$, $\mathcal{C}^\mathrm{cascade}_\mathrm{co}$ cascades
$\mathcal{C}_\mathrm{denoise}$ and $\mathcal{C}_\mathrm{domain}$ per step.
%Et=(t, NLM, Dt) =(t, NLM, (t, NMT, DXYbackground))                          (7) 
 
%And then Formula (1) is redefined into (2) by using E_t.
 
                          %   Wt(x, y)={0, otherwise1|Et| , if (x,y)  Et         (8)
 %Toy example.
 
 \begin{comment}
 \begin{table}[t]
\small
\centering
  %\begin{tabularx}{\textwidth}{l|lll}
  \begin{tabularx}{\linewidth}{l|lll}
{\bf Curriculum} &  {\bf $W_1$ } & {\bf $W_2$} & {\bf $W_3$} \\
\hline 
\makecell[l]{Domain curriculum   \\  $(3)\rightarrow(2)\rightarrow(1)$} &
    $\begin{pmatrix}1/3 \\ 1/3 \\ 1/3\end{pmatrix}$ &
    $\begin{pmatrix}1/2 \\ 1/2 \\ 0.0 \end{pmatrix}$ &
    $\begin{pmatrix}1.0 \\ 0.0 \\ 0.0 \end{pmatrix}$  \\
\makecell[l]{Denoising curriculum  \\ $(2)\rightarrow(1)\rightarrow(3)$} &
    $\begin{pmatrix}1/3 \\ 1/3 \\ 1/3\end{pmatrix}$ &
    $\begin{pmatrix}0.0\\ 1/2 \\ 1/2 \end{pmatrix}$ &
    $\begin{pmatrix}0.0 \\ 0.0 \\ 1.0 \end{pmatrix}$  \\
\makecell[l]{Co-curriculum  \\ $(2)\to(1)$ } &
   $ \begin{pmatrix}1/3 \\ 1/3 \\ 1/3\end{pmatrix}$ &
   $ \begin{pmatrix}1/2\\ 0.0 \\ 1/2 \end{pmatrix}$ &
   $ \begin{pmatrix}1.0 \\ 0.0 \\ 0.0 \end{pmatrix}$ 
  \end{tabularx}
\caption{\small Curricula generated from the toy dataset in Table~\ref{CL}, characterized  by the re-weighting, 
$W_t$, over three training steps. \label{curriculum}}
\end{table}
\end{comment}

 At a time step, both pace functions, in their respective paces,  discard examples that become less relevant to their own tasks.
 All surviving examples then have an equal opportunity to be sampled. Even though uniformly sampled, examples that are more relevant are retained
  longer in training and thus weighed more over time. 
  
  Table~\ref{curriculum} shows a toy example of how two curricula are composed.  At step 1, no example is discarded yet, and
 all examples have equal sampling opportunity ($W_1$'s).
  At step 2, the denoising curriculum discards the noisiest example 2, but
  the domain curriculum still keeps all; So  only 1 and 3 are  retained in the co-curriculum ($W_2$). In step 3, 
  the domain curriculum discards the least in-domain example 3, so only 1 is left in the co-curriculum now ($W_3$).
  The denoising curriculum has a slower pace than the domain curriculum.   Over the four steps, example 1 is kept longer thus weighed more.  
 % Note that this composed order is one of the possible orders, determined by the respective paces how the two individual curricular
  %discard low-quality examples.

  %%%%% and closer to the end.
  %%%% that are further towards the later training.
 %When we decide if an example can be sampled (having non-zero sampling weight), it needs to first belong to
% $\beta(t)$-th percentile, Dt, of the entire data, according to NMT(x,y) and then check if it belongs to the  (t)-th percentile of Dt according to %the scoring function  NLM(x,y).     
      
%// Proof that this is a curriculum per (Bengio et al, 2017)
%// the difference from the AVG.
 
%Both curricula generators utilize the denoising scoring function and the query domain selection function, in different ways. A curricula generator is used in every iteration by an iterative EM optimization algorithm.

\subsection{Curriculum Optimization}

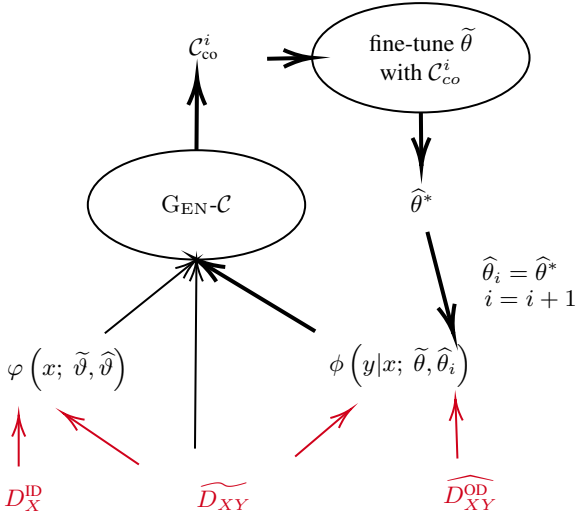
\begin{figure}[t]
\small
\tikzset{every picture/.style={line width=0.75pt}} %set default line width to 0.75pt        

\begin{tikzpicture}[x=0.75pt,y=0.75pt,yscale=-1,xscale=1]
%un if require: \path (0,722.921875); %set diagram left start at 0, and has height of 722.921875

%Shape: Ellipse [id:dp7724959360013537] 
\draw   (54,102.42) .. controls (54,87.23) and (78.29,74.92) .. (108.25,74.92) .. controls (138.21,74.92) and (162.5,87.23) .. (162.5,102.42) .. controls (162.5,117.61) and (138.21,129.92) .. (108.25,129.92) .. controls (78.29,129.92) and (54,117.61) .. (54,102.42) -- cycle ;
%Straight Lines [id:da6201850833194105] 
\draw    (62.5,167) -- (106.7,131.18) ;
\draw [shift={(108.25,129.92)}, rotate = 500.98] [color={rgb, 255:red, 0; green, 0; blue, 0 }  ][line width=0.75]    (10.93,-3.29) .. controls (6.95,-1.4) and (3.31,-0.3) .. (0,0) .. controls (3.31,0.3) and (6.95,1.4) .. (10.93,3.29)   ;

%Straight Lines [id:da3593341116707589] 
\draw [line width=0.75]    (107.5,225) -- (108.23,131.92) ;
\draw [shift={(108.25,129.92)}, rotate = 450.45] [color={rgb, 255:red, 0; green, 0; blue, 0 }  ][line width=0.75]    (10.93,-3.29) .. controls (6.95,-1.4) and (3.31,-0.3) .. (0,0) .. controls (3.31,0.3) and (6.95,1.4) .. (10.93,3.29)   ;

%Straight Lines [id:da05130383749835232] 
\draw [line width=1.5]    (108.25,74.92) -- (108.48,40.92) ;
\draw [shift={(108.5,37.92)}, rotate = 450.39] [color={rgb, 255:red, 0; green, 0; blue, 0 }  ][line width=1.5]    (14.21,-4.28) .. controls (9.04,-1.82) and (4.3,-0.39) .. (0,0) .. controls (4.3,0.39) and (9.04,1.82) .. (14.21,4.28)   ;

%Shape: Ellipse [id:dp7417796753470836] 
\draw   (166,28.42) .. controls (166,13.23) and (190.29,0.92) .. (220.25,0.92) .. controls (250.21,0.92) and (274.5,13.23) .. (274.5,28.42) .. controls (274.5,43.61) and (250.21,55.92) .. (220.25,55.92) .. controls (190.29,55.92) and (166,43.61) .. (166,28.42) -- cycle ;
%Straight Lines [id:da15981621676216107] 
\draw [line width=1.5]    (143.5,28.92) -- (163,28.49) ;
\draw [shift={(166,28.42)}, rotate = 538.73] [color={rgb, 255:red, 0; green, 0; blue, 0 }  ][line width=1.5]    (14.21,-4.28) .. controls (9.04,-1.82) and (4.3,-0.39) .. (0,0) .. controls (4.3,0.39) and (9.04,1.82) .. (14.21,4.28)   ;

%Straight Lines [id:da570621612992869] 
\draw [line width=1.5]    (220.25,55.92) -- (220.47,79.92) ;
\draw [shift={(220.5,82.92)}, rotate = 269.47] [color={rgb, 255:red, 0; green, 0; blue, 0 }  ][line width=1.5]    (14.21,-4.28) .. controls (9.04,-1.82) and (4.3,-0.39) .. (0,0) .. controls (4.3,0.39) and (9.04,1.82) .. (14.21,4.28)   ;

%Straight Lines [id:da503835109075212] 
\draw [line width=1.5]    (167.5,166) -- (110.81,131.48) ;
\draw [shift={(108.25,129.92)}, rotate = 391.34000000000003] [color={rgb, 255:red, 0; green, 0; blue, 0 }  ][line width=1.5]    (14.21,-4.28) .. controls (9.04,-1.82) and (4.3,-0.39) .. (0,0) .. controls (4.3,0.39) and (9.04,1.82) .. (14.21,4.28)   ;

%Straight Lines [id:da0977669657543927] 
\draw [line width=1.5]    (223.5,116) -- (235.76,164.09) ;
\draw [shift={(236.5,167)}, rotate = 255.7] [color={rgb, 255:red, 0; green, 0; blue, 0 }  ][line width=1.5]    (14.21,-4.28) .. controls (9.04,-1.82) and (4.3,-0.39) .. (0,0) .. controls (4.3,0.39) and (9.04,1.82) .. (14.21,4.28)   ;

%Straight Lines [id:da19049747681371065] 
\draw [color={rgb, 255:red, 208; green, 2; blue, 27 }  ,draw opacity=1 ]   (19.5,234) -- (19.5,210) ;
\draw [shift={(19.5,208)}, rotate = 450] [color={rgb, 255:red, 208; green, 2; blue, 27 }  ,draw opacity=1 ][line width=0.75]    (10.93,-3.29) .. controls (6.95,-1.4) and (3.31,-0.3) .. (0,0) .. controls (3.31,0.3) and (6.95,1.4) .. (10.93,3.29)   ;

%Straight Lines [id:da8366841284162836] 
\draw [color={rgb, 255:red, 208; green, 2; blue, 27 }  ,draw opacity=1 ]   (238.5,230) -- (237.57,202) ;
\draw [shift={(237.5,200)}, rotate = 448.09] [color={rgb, 255:red, 208; green, 2; blue, 27 }  ,draw opacity=1 ][line width=0.75]    (10.93,-3.29) .. controls (6.95,-1.4) and (3.31,-0.3) .. (0,0) .. controls (3.31,0.3) and (6.95,1.4) .. (10.93,3.29)   ;

%Straight Lines [id:da3567151712951828] 
\draw [color={rgb, 255:red, 208; green, 2; blue, 27 }  ,draw opacity=1 ]   (79.5,233) -- (43.15,208.13) ;
\draw [shift={(41.5,207)}, rotate = 394.38] [color={rgb, 255:red, 208; green, 2; blue, 27 }  ,draw opacity=1 ][line width=0.75]    (10.93,-3.29) .. controls (6.95,-1.4) and (3.31,-0.3) .. (0,0) .. controls (3.31,0.3) and (6.95,1.4) .. (10.93,3.29)   ;

%Straight Lines [id:da884956946292621] 
\draw [color={rgb, 255:red, 208; green, 2; blue, 27 }  ,draw opacity=1 ]   (157.5,229) -- (183.98,206.3) ;
\draw [shift={(185.5,205)}, rotate = 499.4] [color={rgb, 255:red, 208; green, 2; blue, 27 }  ,draw opacity=1 ][line width=0.75]    (10.93,-3.29) .. controls (6.95,-1.4) and (3.31,-0.3) .. (0,0) .. controls (3.31,0.3) and (6.95,1.4) .. (10.93,3.29)   ;

% Text Node
\draw (721,21) node   {$0$};
% Text Node
\draw (701,71) node   {$0$};
% Text Node
\draw (122,249) node [color={rgb, 255:red, 208; green, 2; blue, 27 }  ,opacity=1 ]  {$\widetilde{D_{XY}}$};
% Text Node
\draw (44,185) node   {$\varphi \left( x;\ \widetilde{\vartheta } ,\widehat{\vartheta }\right)$};
% Text Node
\draw (108.25,102.42) node   {$\mathrm{G}_\mathrm{EN}\text{-}\mathcal{C}$};
% Text Node
\draw (112,23) node   {$\mathcal{C}_{\text{co}}^{i}$};
% Text Node
\draw (220.25,28.42) node   {\makecell{ fine-tune  $\widetilde{\theta}$ \\ $\text{with } \mathcal{C}_{co}^i$}};
% Text Node
\draw (221,100) node   {$ \widehat{\theta}^\ast $};
\draw (270,135) node   {$ \widehat{\theta}_i = \widehat{\theta}^\ast $};
\draw (275,150) node   {$ i = i+1 $};
% Text Node
\draw (210,183) node   {$\phi \left( y|x;\ \widetilde{\theta } ,\widehat{\theta}_i\right)$};
% Text Node
\draw (22,250) node [color={rgb, 255:red, 208; green, 2; blue, 27 }  ,opacity=1 ]  {$D^{\text{ID}}_{X}$};
% Text Node
\draw (245,247) node [color={rgb, 255:red, 208; green, 2; blue, 27 }  ,opacity=1 ]  {$\widehat{D^{\text{OD}}_{XY}}$};

\end{tikzpicture}

\caption{
\footnotesize
Co-curricular learning with an EM-style optimization procedure. Thicker arrows form the bootstrapping  loop. 
\label{cocurricular_fig}
}
\end{figure}

We further improve the co-curriculum using an EM \cite{Dempster77maximumlikelihood} style optimization procedure in training,
as shown in Figure~\ref{cocurricular_fig}. It aims specifically to iteratively improve
the denoising selection, without losing quality on the domain selection.

With $\widetilde{D_{XY}}$ and $D_{X}^\mathrm{ID}$, we train a domain scoring function, $\varphi(x; \widetilde{\vartheta}, \widehat{\vartheta})$.
With  $\widetilde{D_{XY}}$ and $\widehat{D_{XY}^{\mathrm{OD}}}$, we train a denoising scoring function,
$\phi(y|x; \widetilde{\theta}, \widehat{\theta})$. The in-domain component $\widehat{\vartheta}$ 
of $\varphi$ or the clean component $\widehat{\vartheta}$ of $\phi$ are obtained by fine-tuning $\widetilde{\vartheta}$ or
$\widetilde{\theta}$ on the respective seed data. These {\em initialize} the procedure (iteration 0).

At iteration $i$, we generate a concrete co-curriculum using the dynamic re-weighting, $W_t$,
as  defined in Section~\ref{our_approach}.
Let G{\small EN}-$\mathcal{C}$ denote the {\em curriculum generation} process:
\begin{equation}
\mathcal{C}_\mathrm{co} = \mbox{G{\small EN}-}\mathcal{C}\left(\widetilde{D_{XY}}, \phi_i, \varphi\right) \label{e-step}
\end{equation}

Then, we fine-tune the original noisy NMT component, $\widetilde{\theta}$, of $\phi$  on $\mathcal{C}_\mathrm{co}$:
\begin{equation}
\widehat{\theta}^{*} = \arg\max_{\widehat{\theta}} \mathcal{P} \left(\mathcal{C}_\mathrm{co}, m_f\right)
\end{equation}
 $\widehat{\theta}^{*}$ is used to replace the  clean component of  $\phi$
 \begin{eqnarray}
\widehat{\theta}_i &=& \widehat{\theta}^{\ast} \nonumber \\
i &=& i+1 \nonumber
\end{eqnarray}
$\widehat{\theta}_i$ is then compared against the original  $\widetilde{\theta}$  for scoring.
The updated $\phi$ and the constant $\varphi$ work together to generate a new co-curriculum in the next
iteration going back to Eq.~\ref{e-step}. In this process, only the denoising function $\phi$ is iteratively updated, made
more aware of the domain. 
%$\widehat{\theta}$ and $\mathcal{C}_\mathrm{co}$ bootstrap each other.
%and $\widehat{\theta}$ bootstrap each other. 
%The process
%finishes after a pre-defined number of iterations. This is the {\em optimization step}. 

We call the procedure EM-style because $\widetilde{D_{XY}}$ is treated as incomplete
without the (hidden) data order. The generated $\mathcal{C}_\mathrm{co}$ in each iteration sorts the data and thus
is viewed as complete. It is then used to train $\widehat{\theta}$ by maximizing the performance of the final task.
% using  $\mathcal{C}_\mathrm{co}$.
$\widehat{\theta}$ and $\mathcal{C}_\mathrm{co}$  bootstrap each other.
The process finishes after a pre-defined number of iterations. We use shallow parameterization for scoring functions but we can
train a deep model  on the final $\mathcal{C}_\mathrm{co}$. The process also uses fine-tuning, so it can be run efficiently.

In principle, the domain-data scoring function $\varphi$ can be updated in a similar manner, too, by updating its in-domain component,
$\widehat{\vartheta}$. This may help when the in-domain monolingual corpus is very small.
 An alternating optimization process can be used to bootstrap both.  We, however, do not investigate this.

\section{Experiments}
%We run experiments to examine the impact of co-curricular learning for the problem stated in Section~\ref{problem}. 
%Particularly, we see how the method can meet objectives in Eq~\ref{p1},~\ref{p2}
% and~\ref{p3}, so gain understanding about transfer learning of data quality.

\subsection{Setup} \label{setup}
We consider two  background datasets and two test domains, so we have four experiment configurations.
Each configuration has as inputs a background dataset,  an in-domain source-language corpus and a (small) trusted parallel
dataset that is  out-of-domain. The inputs of a configuration are shown in Figure~\ref{cocurricular_fig}.

As alternative background datasets, we use the English$\to$French Paracrawl data,\footnote{
\url{https://paracrawl.eu}} (300 million pairs),
 and the WMT14 training data (40 million  pairs).
The former is severely noisier than the later.
We adopt sentence-piece model and apply open-source implementation  \cite{P18-1007} to segment 
 data into sub-word units with a source-target shared 32000 sub-word vocabulary.

We use two test domains: the English$\to$French IWSLT15 test set, in spoken language domain; 
and the English$\to$French WMT14 test set, in news domain. 
 For  IWSLT15,  we  use the English side of its
 provided parallel training data (220 thousand examples) as $D_X^\mathrm{ID}$,  but use the parallel
 version as $\widehat{D_{XY}^\mathrm{OD}}$ for the WMT14 domain.  The IWSLT14 test set is
 used for validation.
 For  the WMT14 domain,  the provided  28 million English sentences are used as $D_X^\mathrm{ID}$. WMT 2010-2011 test sets
  are concatenated as $\widehat{D_{XY}^\mathrm{ID}}$ for news\footnote{
 Strictly speaking, though all are in news, the WMT 2014 monolingual data,  the WMT 2011-2012 test sets and the 2014 test set
  are not necessarily in the exact same news domain. So this news test domain could be treated as a looser case than the 
  IWSLT domain and examines the method at a slightly different position in the spectrum of the problem.
   }, or as $\widehat{D_{XY}^\mathrm{OD}}$ for the above IWSLT15 test domain.
 So, the trusted data are reversely shared across the two  test domains.  Additionally, WMT 2012-2013 are 
 used  as the validation set for the WMT14 test domain.
   Our method does not
 require the in-domain trusted data, but we use it to construct bounds in evaluation.

%The  IWSLT15 training data is used as $\widehat{D}_{\text{enfr}}^{\text{\small OD}}$, the configuration of
%which is merely the revert of above IWSLT15 test case.  WMT newstest2011-12 are used as the validation set  and
 %newstest2014 as the test set. We also removed test overlap  from training.
 
%in-domain, monolingual queries $D_{\text{en}}^{\text{\small ID}}$. There are 220 thousand sentences.
% WMT newstest10-11 are used as the out-of-domain, trusted data $\widehat{D}_{\text{enfr}}^{\text{\small OD}}$.
  %IWSLT14 test set is ftion set for training models and IWSLT15  as the test set. 

We use RNN-based NMT  \citep{wu2016}   to train models. Model parameterization for
 $\theta$'s  of $\phi$ (Eq~\ref{phi}) or $\vartheta$'s $\varphi$ (Eq~\ref{varphi}) is 512 dimensions by 3 layers -- NLMs are realized using
  NMT models with dummy  source sentences \cite{sennrich-haddow-birch:2016:P16-11}.  Deep models  are 1024 dimensions by 8 layers. 
  Unless specified, results are reported for deep models.
    We compute truecased, detokenized \BLEU with {\tt mteval-v14.pl}.

Training on Paracrawl uses Adam in warmup and then SGD for a total of 3 million steps using batch size 128, learning
rate 0.5 annealed, at step 2 million, down to 0.05. 
Training on WMT 2014 uses batch size 96, dropout probability 0.2 for a total of 2 million steps, with learning rate 0.5 annealed, at 
step 1.2 million, down to 0.05, too.
No dropout is used in Paracrawl training due to its large data volume. 
%We use same learning rate scheduls (before and after the annealing) between the two data sets?

For the pace hyper-parameters (Section~\ref{our_approach}), we empirically use $H=F=400k$, $G=900k$. Floor values set for
$\lambda, \beta, \gamma$ are top $0.1, 0.2, 0.5$ selection ratios, respectively, such that in the cascaded co-curriculum case, the  tightest effective 
percentile value would be the same $0.1=0.2 \times 0.5$, too. All single curriculum experiments use the same pace setting
as $\mathcal{C}^\mathrm{mix}$.

\subsection{Baselines and Oracles}

 We build various systems  below as baselines and oracles. Oracle systems use in-domain trusted 
 parallel data.
 %($\widehat{D}_{\text{enfr}}^{\text{\small ID}}$).
 %represent existing best practices of using
 % in-domain parallel data and are used for us to assess where we stand after quality transfer. The oracles are
 % reachable, when in-domain parallel corpus is available.

\vspace{-0.1in}

\begin{itemize}[leftmargin=*]
\item[] \underline{Baselines}:
\begin{itemize}[leftmargin=*]
  \item[1.]  $\mathcal{C}_{\text {random}}$ :  Baseline model trained  on background data with random data sampling.
%The proposed co-curricular learning method works as a solution when $\widehat{D}_{\text{enfr}}^{\text{\small ID}}$ is unavailable. Thus, this system serves as a kind of upper-bound reference.
 
  \item[2.] $\mathcal{C}_\text{domain}$: Dynamically fine-tunes $\mathcal{C}_{\text {random}}$ with a domain curriculum \cite{dynamiccds}.
  
\item[3.]  $\mathcal{C}_{\text{denoise}}$: Dynamically fine-tunes $\mathcal{C}_{\text {random}}$ with a denoising curriculum \cite{denoise_nmt}.

 \end{itemize}
\item[] \underline{Oracles}:
  \begin{itemize}[leftmargin=*]
     \item[4.] $\mathcal{C}_{\text{true}}$: 
   Dynamically fine-tunes $\mathcal{C}_{\text {random}}$ with the true curriculum.
   
     \item[5.] ID fine-tune $\mathcal{C}_{\text {random}}$: Simply fine-tunes  $\mathcal{C}_{\text {random}}$  with   in-domain (ID) parallel data.

  %  \item[6.] Co-curriclum + ID fine-tune: ID fine-tune the model trained on final co-curriclum.

  \end{itemize}

%\item  ID fine tune of 5: Further fine tunes 5 with in-domain bilingual trusted data,  expected to perform  even stronger 
% than 5.
 
 \end{itemize}
 We'll see if our method is better than either original curriculum and how close it is to the true curriculum oracle.
 In most experiments, we fine-tune a warmed-up (baseline) model to compare curricula, for
 quicker experiment cycles.
    
    %Depending on the diversity of $\widetilde{D}_{\text{\small enfr}}^{\text{background}}$, our method may reach 2 and 6 or fall
  %behind -- A  parallel dataset may always contain sentence pairs that exactly match the query domain. 
   
   \begin{table}[]
\small
\begin{center}
\begin{tabular}{l|cc}
%\hline
\multirow{2}{*}{\bf Models} & \multicolumn{2}{c}{{\bf Test \BLEU} } \\ %\cline{2-3} 
                         & {\bf IWSLT15}     & {\bf WMT14 }    \\ %\cline{1-1}
\hline\hline

%\multicolumn{3}{l}{} \\
\multicolumn{3}{l}{(P)aracrawl} \\
%\hdashline
%\multicolumn{3}{l}{} \\
P1: $\mathcal{C}_{\text{random}}$            &      34.6    &     31.6               \\
P2: $\mathcal{C}_\text{domain}$   &     35.7        &     32.4               \\ % \hline
P3: $\mathcal{C}_{\text{denoise}}$ & 36.6  & 33.6 \\
%\hdashline
P4: $\mathcal{C}_{\text{true}}$ & 37.2 & 34.2 \\
P5: ID fine-tune P1                          &      38.5       &   34.0                 \\
% P/R6: ID fine-tune of P/B5  & & \\
\hline
%\multicolumn{3}{l}{} \\
\multicolumn{3}{l}{(W)MT} \\
%\hdashline
%\multicolumn{3}{l}{} \\
W1: $\mathcal{C}_{\text{random}}$            &    36.5      &     35.0               \\
W2: $\mathcal{C}_\text{domain}$   &     37.6       &     35.9              \\ % \hline
W3: $\mathcal{C}_{\text{denoise}}$ & 37.4  & 36.0 \\ 
%\hdashline
W4: $\mathcal{C}_{\text{true}}$ & 38.5  & 36.3 \\
W5: ID fine-tune W1                          &      39.7       &   35.9                 \\

% W/R6: ID fine-tune W/B5   & & \\

\end{tabular}
\end{center}
\caption{\small Baseline and oracle models trained on Paracrawl data and WMT data, respectively. ID: in-domain. P2,3,4 (or W2,3,4) 
each dynamically fine-tunes P1 (or W1) with the respective curriculum.  Except for P1 and W1, the two \BLEU scores in each row are for
{\em two different}  training runs, each focusing on its own test domain (configuration).
\label{baselines}}
\end{table}

Baseline and oracle \BLEU scores are shown in Table~\ref{baselines}.  Note that, except for P1 and W1, the two \BLEU scores in a row are for
{\em two different}  training runs, each focusing on its own test domain.
On either training dataset, domain curriculum, $\mathcal{C}_\mathrm{domain}$,
improves baseline, $\mathcal{C}_\mathrm{random}$,  by 0.8-1.1 \BLEU (P3 vs P1, W3 vs W1).  
$\mathcal{C}_\mathrm{domain}$ falls behind of $\mathcal{C}_\mathrm{denoise}$ on the noisy Paracrawl
dataset (P2 vs P3), but delivers matched performance on the cleaner WMT dataset (W2 vs W3) -- noise compromises the domain capability.
On the WMT training data, $\mathcal{C}_\mathrm{denoise}$ improves baselines by about +1.0 \BLEU on either test domain (W3 vs W1), and more on the noisier Paracrawl data: +2.0 on either test domain (P3 vs P1).
The true curriculum (P4, W4) bounds the performance of $\mathcal{C}_\mathrm{domain}$ and $\mathcal{C}_\mathrm{denoise}$. Simple in-domain fine-tuning gives good improvements  (P5 vs P1, W5 vs W1). 

%In-domain trusted parallel data gives the best results when directly used to fine-tune the baseline system (P5 vs P1).

% Systems (2, 3, 5, 6) represent some of those best practices in published literatures for pushing a model's performance with trusted data, either in-domain or out-of-domain.
 
 %  According to requirements specified in Formulae~\ref{p1}~\ref{p2}~and~\ref{p3}, our proposed method aims to
%   work better than 1, 3, 4 and as close as to 5, by using only a query corpus and some out-of-domain curated data.

 %[comment on the strength of the WMT system later]
 \subsection{Co-Curricular Learning}
 
%\subsection*{Addition vs. composition}
{\bf Cascading vs. mixing}.  Table~\ref{avg_vs_nested} shows per-step 
cascaded filtering can work better than flat mixing (P7 vs P6). 
 So we  use 
 $\mathcal{C}_{\text{co}}^{\text{\small cascade}}$ for the remaining experiments.
 \vspace{0.05in}
 
%\vspace{0.1in}

\begin{table}[h]
\small
\begin{center}
\begin{tabular}{l|cc}
%\hline
\multirow{2}{*}{{\bf \makecell[l]{Co-\\Curriculum}}} & \multicolumn{2}{c}{{\bf Test \BLEU}}  \\ %\cline{2-3} 
                         & {\bf IWSLT15 }     & {\bf WMT14 }    \\ %\cline{1-1}
\hline\hline
P6: $\mathcal{C}_{\text{co}}^{\text{mix}} $            &    36.2      &     33.8              \\
P7: $\mathcal{C}_{\text{co}}^{\text{cascade}} $            & {\bf 37.1}         &  {\bf 34.0}                   \\
\end{tabular}
\end{center}
\caption{\small
Per-step cascading works better than mixing on Paracrawl data.
\label{avg_vs_nested}
}
\end{table}

%\subsection{\BLEUs of co-curricular learning}

\noindent
{\bf Curriculum \BLEU comparisons}. Table~\ref{co-CL-bleus} shows the effectiveness of
co-curricular learning. On Paracrawl, co-curriculum (P7) gives more than +2 \BLEU
on top of no CL (P1). It improves $\mathcal{C}_\mathrm{domain}$ (P7 vs P2) by +1.4 \BLEU
on IWSLT15 and +1.6 \BLEU on WMT14. It is better than either constituent curriculum (P2 or P3),
close to the true curriculum (P4).

   \begin{table}[t]
\small
\begin{center}
\begin{tabular}{l|cc}
%\hline
\multirow{2}{*}{{\bf Curriculum}} & \multicolumn{2}{c}{{\bf Test \BLEU}}  \\ %\cline{2-3} 
                         & {\bf IWSLT15 }     & {\bf WMT14 }    \\ %\cline{1-1}
\hline\hline
P1: $\mathcal{C}_{\text{random}}$            &      34.6    &     31.6               \\
P2: $\mathcal{C}_\text{domain}$   &     35.7        &     32.4               \\ % \hline
P3: $\mathcal{C}_{\text{denoise}}$ & 36.6  & 33.6 \\
P7: $\mathcal{C}_\mathrm{co}$ & {\bf 37.1} & {\bf 34.0} \\
$\mathcal{C}_\mathrm{co} - \mathcal{C}_\mathrm{domain}$    &  ${\it +1.4}$ & ${\it +1.6}$  \\
 $\mathcal{C}_\mathrm{co} - \mathcal{C}_\mathrm{true}$    &  ${\it -0.1}$ & ${\it -0.2}$  \\
%\makecell[l]{P7: $\mathcal{C}_\mathrm{co}$ \\ $^{}$ \\ } & \makecell[c]{{\bf 37.1} \\ $^{\it +1.4}$ \\ $^{\it -0.1}$ }
%      & \makecell[c]{{\bf 34.0} \\ $^{\it 1.6}$ \\ $^{\it -0.2}$ } \\
\hline
W1: $\mathcal{C}_{\text{random}}$            &    36.5      &     35.0               \\
W2: $\mathcal{C}_\text{domain}$   &     37.6       &     35.9              \\ % \hline
W3: $\mathcal{C}_{\text{denoise}}$ & 37.4  & 36.0 \\ 
W7:  $\mathcal{C}_\mathrm{co}$  & {\bf 37.8}  & {\bf 36.4} \\
$\mathcal{C}_\mathrm{co} - \mathcal{C}_\mathrm{domain}$    &  ${\it +0.2}$ & ${\it +0.5}$  \\
$\mathcal{C}_\mathrm{co} - \mathcal{C}_\mathrm{true}$    &  ${\it -0.7}$ & ${\it +0.1}$  \\
\end{tabular}
\end{center}
\caption{\small
Co-curriculum improves either constituent curriculum and no CL, can be close to the true
curriculum on noisy data.
%Scores in italics are improvement from  $\mathcal{C}_\mathrm{co}$ over  $\mathcal{C}_\mathrm{domain}$.
\label{co-CL-bleus}
}
\end{table}

On the cleaner WMT training data, co-curriculum (W7)  improves 
 either constituent curricula (W2 and W3) by smaller gains  than Paracrawl: +0.2
 \BLEU on IWSLT15  and +0.4 on WMT14. Compared to $\mathcal{C}_\mathrm{true}$ W5,
 co-curriculum W7 falls behind  (-0.7 \BLEU) on IWSLT15 and matches (+0.1 \BLEU) on WMT14.

 So  $\mathcal{C}_\mathrm{co}$  outperforms either constituent curriculum, as we
 target in Section~\ref{problem}.
 %It seems to be more effective when
% the background data is noisy, in which case, its performance seems closer to the true curriculum in our case.
 In both background data cases,  using in-domain trusted parallel data to build oracles (P5, W5) are  more effective than selecting data
 in our setup.

\begin{comment}
 \begin{table}[t]
\small
\begin{center}
\begin{tabular}{l|cc}
%\hline
\multirow{2}{*}{{\bf \makecell[l]{Model}}} & \multicolumn{2}{c}{{\bf Test \BLEU}}  \\ %\cline{2-3} 
                         & {\bf IWSLT15 }     & {\bf WMT14 }    \\ %\cline{1-1}
\hline\hline
\multicolumn{3}{l}{$D_{\mathrm{enfr}}^{\mathrm{background}}=\mathrm{(P)aracrawl}$} \\
\hdashline
%\multicolumn{3}{l}{} \\
             \\ % \hline
P7: $\mathcal{C}_\mathrm{co}$ & 37.1 & 34.0 \\
P8: Optimized $\mathcal{C}_{\mathrm{co}}$ & 37.3 & 34.6 \\
P8 - $\mathcal{C}_\mathrm{domain}$    &  ${\it +1.6}$ & ${\it +2.0}$  \\
\hdashline
P9: Retrain on optimized $\mathcal{C}_\mathrm{co}$ & {\bf 37.9} &   {\bf 35.6}  \\
%P12: ID fine-tune P9 & & \\
\hdashline
P10: Fine-tune on static sel. & 36.8 & 34.6 \\
P11: Retrain on static sel. & 37.1 & 34.6 \\
\hline
\multicolumn{3}{l}{$D_{\mathrm{enfr}}^{\mathrm{background}}=\mathrm{(W)MT}$} \\
\hdashline
%\multicolumn{3}{l}{} \\
W7: $\mathcal{C}_\mathrm{co}$ & 37.8 & 36.4 \\
W8: Optimized $\mathcal{C}_\mathrm{co}$  & 37.8 & {\bf 36.5} \\
\hdashline
W9: Retrain on  optimized $\mathcal{C}_\mathrm{co}$  & {\bf 38.1 } & 36.3 \\
%W12: ID fine-tune W9 & 39.9  & 36.9 \\
\hdashline
W10: Fine-tune on static sel. & 37.4 & 36.2 \\
W11: Retrain on static sel. & 34.0 & 31.7 \\
\end{tabular}
\end{center}
\caption{\small
{\BLEU}s of co-curricular learning. ``Optimized'': Optimize co-curriculum by EM-style optimization. The static selection experiments use top 10\% 
selection. Bold font marks the best curriculum learning performance.
\label{BLEU}
}
\end{table}
\end{comment}

\subsection{Effect of Curriculum Optimization} \label{EM_effect}

We further bootstrap the co-curriculum with the EM-style optimization procedure (Figure~\ref{cocurricular_fig})  for three iterations for all four configurations.
%And examine the impact on the final translation model.
% In each iteration, the re-generated co-curriculum dynamically fine-tunes the original  scoring component $\widetilde{\theta}$ to yield a new $\widehat{\theta}$.
  \vspace{0.05in}

\begin{figure}[t]
\begin{center}
\begin{tikzpicture}[scale=0.6]
\begin{axis}[ybar,
             xlabel={Iteration}, ylabel={\BLEU},
             ymax=35, ymin=22, minor y tick num = 0,
             bar width=7pt, 
             xtick=data, 
             symbolic x coords={baseline, clean, EM-$1$, EM-$2$, EM-$3$}, 
             legend pos=north west],
\addplot coordinates { (baseline, 23.9 ) (clean, 28) (EM-$1$, 32.8) (EM-$2$, 34.1) (EM-$3$, 34.1) };
\addplot coordinates { (baseline, 23.3 ) (clean, 24.7) (EM-$1$, 29.2) (EM-$2$, 30.7) (EM-$3$, 31.5) };
\legend{IWSLT15, WMT14}
\end{axis}
\end{tikzpicture}
\end{center}
\caption{\small The EM-style optimization has a big impact on  small-capacity models, measured in \BLEU. Experiments were carried out on Paracrawl data. 
%Connecting to Figure~\ref{cocurricular_fig}, x-axis label `baseline': $\widetilde{\theta}$; label `clean': clean model $\widehat{\theta}$ by fine-tuning $\widetilde{\theta}$ on clean data;  `EM-$i$': bootstrapped $\widehat{\theta}$ 
%per iteration.  
}
    \label{bleu_over_em_iterations}
\end{figure}
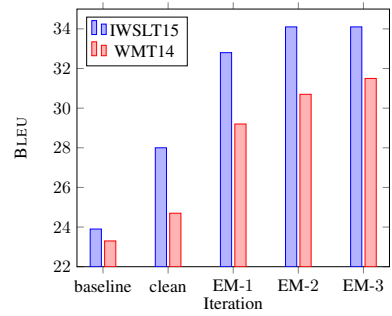

\noindent
{\bf Shallow models}. We  use the  translation performance of the clean component $P(y|x; \widehat{\theta})$ in
scoring function $\phi$ (Eq.~\ref{phi}) as  an indicator to the quality of  $\mathcal{C}_{\text{co}}$ per iteration. 
Figure~\ref{bleu_over_em_iterations} shows that the  \BLEU scores of  $P(y|x; \widehat{\theta})$ steadily become better by iterations.\footnote{
They also include  two initialization points: the noisy $\widetilde{\theta}$, and the initial clean
$\widehat{\theta}$ obtained by fine-tuning $\widetilde{\theta}$ on the clean data.
}  $\widehat{\theta}$ has 512 dimensions and 3 layers. Surprisingly, EM-$3$ improves baseline by +10 BLEU on IWSLT15, +8.2 BLEU on WMT14 and performs better than fine-tuning baseline with the clean, out-of-domain parallel data we have. They even reach the performance of $\mathcal{C}_\mathrm{random}$ (P1) that uses a much deeper model (1024 dimensions x 8 layers) trained on the vanilla data.
%Optimization
%improves co-curriculum to train shallower models (512 dimensions by 3 layers).
\vspace{0.05in}

\noindent
{\bf Deep models}. Table~\ref{EM-bleus} shows the {\BLEU}s of deep models (1024 dimensions x 8 layers) trained on
the final co-curriculum. 
 P8 performs slightly better than the non-bootstrapped version P7 on Paracrawl: +0.6 BLEU on WMT14 test and +0.2 on IWSLT15 test.
The differences on the WMT data appear to be smaller (W8 vs. W7).
So, curriculum bootstrapping has a  small impact overall on deep models.

\begin{table}[t]
\small
\begin{center}
\begin{tabular}{l|cc}
%\hline
\multirow{2}{*}{{\bf Curriculum}} & \multicolumn{2}{c}{{\bf Test \BLEU}}  \\ %\cline{2-3} 
                         & {\bf IWSLT15 }     & {\bf WMT14 }    \\ %\cline{1-1}
\hline\hline
P2: $\mathcal{C}_\text{domain}$   &     35.7        &     32.4  \\ 
P7: $\mathcal{C}_\mathrm{co}$ &  37.1  & 34.0 \\
P8: P7+Optimization & {\bf 37.3} & {\bf 34.6} \\
P8 - $\mathcal{C}_\mathrm{domain}$    &  ${\it +1.6}$ & ${\it +2.0}$  \\
\hline
W2: $\mathcal{C}_\text{domain}$   &     37.6       &     35.9              \\ 
W7: $\mathcal{C}_\mathrm{co}$ & {\bf 37.8} &  36.4 \\
W8: W7+Optimization  & {\bf 37.8} & {\bf 36.5} \\
W8 - $\mathcal{C}_\mathrm{domain}$    &  ${\it +0.2}$ & ${\it +0.6}$  \\
\end{tabular}
\end{center}
\caption{\small
EM-style optimization further improves domain curriculum. But, overall, it has a small impact on deep models.
\label{EM-bleus}
}
\end{table}

\paragraph{Why the difference?}  Why is there such a difference? We analyze the properties of the co-curriculum.

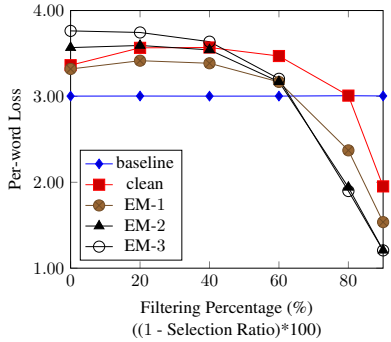
\begin{figure}[t]
\begin{center}
\begin{tikzpicture}[scale=0.6]
\begin{axis}[xlabel={\makecell[c]{Filtering Percentage (\%) \\ (($1$ - Selection Ratio)*100)}}, ylabel={Per-word Loss},
             ymax=4, ymin=1, minor y tick num = 0,
             xmax=90,xmin=0,minor x tick num = 0,
             scaled y ticks = false,
             %cycle list name=black white,
             yticklabel style={%
        /pgf/number format/.cd,
            fixed,
            fixed zerofill,
            precision=2,
            /tikz/.cd
    },
                   legend pos=south west],
             ]
     \addplot+[mark size = 3pt, mark=diamond*] table[x index=0,y index=5,col sep=comma] {perword_loss.dat};
     \addlegendentry{baseline}
    \addplot+[mark size=3.5pt] table[x index=0,y index=1,col sep=comma] {perword_loss.dat};
    \addlegendentry{clean}
    \addplot+[mark size=3.5pt] table[x index=0,y index=2,col sep=comma] {perword_loss.dat};
    \addlegendentry{EM-$1$}
    \addplot+[mark size =3.5pt, mark=triangle*] table[x index=0,y index=3,col sep=comma] {perword_loss.dat};
    \addlegendentry{EM-$2$}
    \addplot+[mark size=3.5pt, mark=o,draw=black] table[x index=0,y index=4,col sep=comma] {perword_loss.dat};
    \addlegendentry{EM-$3$}

%\legend{Paracrawl,WMT}
\end{axis}
\end{tikzpicture}
\end{center}
\caption{\small Curriculum learning and optimization push ``easier-to-learn'' (lower
per-word loss) examples to late curriculum (right) and harder examples (higher per-word
loss) to early curriculum (left). 
%Over iterations/curves, a later optimization iteration schedules
%even easier examples to late curriculum, compared to a previous iteration at the same selection percentile; The story happens reversely at the early phase of curriculum due to probability mass
%conservation.
}
\label{perword-loss}
\end{figure}

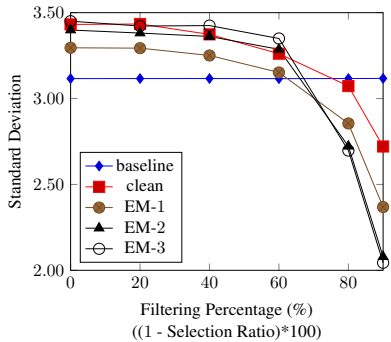
\begin{figure}[t]
\begin{center}
\begin{tikzpicture}[scale=0.6]
\begin{axis}[xlabel={\makecell[c]{Filtering Percentage (\%) \\ (($1$ - Selection Ratio)*100)}}, ylabel={Standard Deviation},
             ymax=3.5, ymin=2.0, minor y tick num = 0,
             xmax=90,xmin=0,minor x tick num = 0,
             scaled y ticks = false,
             %cycle list name=black white,
             yticklabel style={%
        /pgf/number format/.cd,
            fixed,
            fixed zerofill,
            precision=2,
            /tikz/.cd
    },
                   legend pos=south west],
             ]
   
    \addplot+[mark size = 3pt, mark=diamond*] table[x index=0,y index=5,col sep=comma] {SD.dat};
    \addlegendentry{baseline}
    \addplot+[mark size=3.5pt] table[x index=0,y index=1,col sep=comma] {SD.dat};
    \addlegendentry{clean}
    \addplot+[mark size=3.5pt] table[x index=0,y index=2,col sep=comma] {SD.dat};
    \addlegendentry{EM-$1$}
    \addplot+[mark size =3.5pt, mark=triangle*] table[x index=0,y index=3,col sep=comma] {SD.dat};
    \addlegendentry{EM-$2$}
    \addplot+[mark size=3.5pt, mark = o, draw = black] table[x index=0,y index=4,col sep=comma] {SD.dat};
    \addlegendentry{EM-$3$}

%\legend{Paracrawl,WMT}
\end{axis}
\end{tikzpicture}
\end{center}
\caption{\small Curriculum learning and optimization push ``regularized'' (lower
variance) examples to late curriculum  and higher-variance examples
to early curriculum.}
\label{SD}
\end{figure}

Each curve in Figure~\ref{perword-loss} corresponds to a single curriculum
that simulates the online data selection from looser selection
(left x-axis) to more-tightened selection  (right x-axis).  
During the course of a single CL, the curriculum pushes ``harder'' examples with higher per-word loss (than baseline) to the early curriculum
phase (for exploration), and ``easier-to-learn'' examples with lower per-word loss to the late curriculum phase (for exploitation).
Over iterations,  a later-iteration curriculum schedules
even easier examples than a previous iteration at late curriculum. 
The story happens reversely at early curriculum due to probability mass conservation.
Figure~\ref{SD} shows a similar story regarding per-word loss variance.
So, curriculum optimization ``regularizes'' the curriculum and makes it easier-to-learn towards the end of CL.

These may be important
for a small-capacity model to learn efficiently. The fact that the deep model
is not improved as much  means that `clean' may have taken most of the headroom
for deep models.

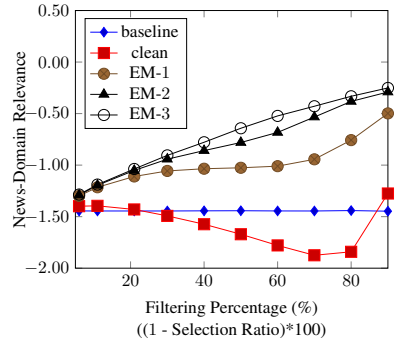
\begin{figure}[t]
\begin{center}
\begin{tikzpicture}[scale=0.6]
\begin{axis}[xlabel={\makecell[c]{Filtering Percentage (\%) \\ (($1$ - Selection Ratio)*100)}}, ylabel={News-Domain Relevance},
             ymax=0.5, ymin=-2, minor y tick num = 0,
             xmax=90,xmin=5,minor x tick num = 0,
             scaled y ticks = false,
             %cycle list name=black white,
             yticklabel style={%
        /pgf/number format/.cd,
            fixed,
            fixed zerofill,
            precision=2,
            /tikz/.cd
    },
                  legend style={fill=none},
                   legend pos=north west],
             ]
   
    \addplot+[mark size = 3pt, mark=diamond*] table[x index=0,y index=1,col sep=comma] {EM-relevance.dat};
    \addlegendentry{baseline}
    \addplot+[mark size=3.5pt] table[x index=0,y index=2,col sep=comma] {EM-relevance.dat};
    \addlegendentry{clean}
    \addplot+[mark size=3.5pt] table[x index=0,y index=3,col sep=comma] {EM-relevance.dat};
    \addlegendentry{EM-$1$}
    \addplot+[mark size =3.5pt, mark=triangle*] table[x index=0,y index=4,col sep=comma] {EM-relevance.dat};
    \addlegendentry{EM-$2$}
    \addplot+[mark size=3.5pt, mark=o, draw = black] table[x index=0,y index=5,col sep=comma] {EM-relevance.dat};
    \addlegendentry{EM-$3$}

%\legend{Paracrawl,WMT}
\end{axis}
\end{tikzpicture}
\end{center}

\caption{\small The denoising curriculum is made more aware of news-domain after iterations. 
Figure drawn for the (Paracrawl, news) configuration.
Within a single curriculum, `baseline' randomly shuffles
data, thus flat curve. `clean' uses the  out-of-domain clean parallel data, thus
not that much news relevance. All curves show negative news-domain relevance, indicating lack
of news data in  Paracrawl data.}
%data may amount only to a small percentage in.}
\label{news-relevance}
\end{figure}

Meanwhile, according to Figure~\ref{news-relevance}, each individual
curriculum concentrates more on news in-domain examples as training progresses.
Over iterations, bootstrapping makes the co-curriculum more news-domain aware. Due to
the use of the denoising curriculum, data in curriculum becomes cleaner, too.
So, although the co-curriculum schedules data from hard to easier-to-learn, which
seems  opposite to the general CL, it also
schedules data from less in-domain to cleaner and more in-domain, which
captures the spirit of CL.

%[This is done with 2M Paracrawl sentence pair sample.]
%[connect to distillation]

% So, overall, the optimization appears to
%bring relatively smaller improvements for deeper models.
% though still may be positive.

%{\bf denoising capability}

\begin{comment}
\begin{figure}[h]
\begin{center}
\includegraphics[width=8.5cm,height=6.5cm]{em_dev_curves.png}
\end{center}

 \caption{
 Perplexity curves of  EM training iterations, measured for  $\widehat{\theta}$, during ``Train NMT'' in Figure~\ref{cocurricular_fig}.
 \label{em_dev_curves}
}
\end{figure}
\end{comment}

%\subsection{BT}
%\subsection{Curves in training}

\subsection{Retraining}

On  Paracrawl, retraining NMT with  co-curriculum improves dynamic fine-tuning, as shown in Table~\ref{retrain-bleus} (P9 vs. P8):
+0.6 \BLEU on IWSLT15  and +1.0 \BLEU on WMT14. 
On  WMT14 training data, retraining   (W9) seems to perform similarly to fine-tuning
on a warmed-up model (W8): 
+0.3 on IWSLT15 but -0.2 on WMT14; We speculate that this may be due to 
 the smaller WMT training data size.
 
  % quality.

  \begin{table}[t]
\small
\begin{center}
\begin{tabular}{l|cc}
%\hline
\multirow{2}{*}{{\bf Curriculum}} & \multicolumn{2}{c}{{\bf Test \BLEU}}  \\ %\cline{2-3} 
                         & {\bf IWSLT15 }     & {\bf WMT14 }    \\ %\cline{1-1}
\hline\hline
P8: Fine-tune with $\mathcal{C}_{\mathrm{co}}$ & 37.3&  34.6 \\
P9: Retrain with $\mathcal{C}_\mathrm{co}$ & {\bf 37.9} &   {\bf 35.6}  \\
\hline
W8: Fine-tune with $\mathcal{C}_\mathrm{co}$  &  37.8 & {\bf 36.5} \\
W9: Retrain with $\mathcal{C}_\mathrm{co}$  & {\bf 38.1 } & 36.3 \\
\end{tabular}
\end{center}
\caption{\small
Retraining with a curriculum may work better than fine-tuning with it, on a large, noisy dataset. 
\label{retrain-bleus}
}
\end{table}

\subsection{Dynamic vs. Static Data Selection}

Co-curricular learning is dynamic. How does being dynamic matter?
Table~\ref{dynamic-vs-static} shows that fine-tuning on the top 10\% data\footnote{
This is the ratio where the pace function
reaches the floor value in training (see end of Section~\ref{setup}).
}
static selection (P10, W10) gives good improvements over baselines P1, W1,  
but  co-curriculum (P9, W9) may do better. This confirms findings by \cite{dynamiccds}.
%More importantly, retraining (P9) on the co-curriculum may further amplify the gains.

\begin{table}[t]
\small
\begin{center}
\begin{tabular}{l|cc}
%\hline
\multirow{2}{*}{{\bf \makecell[l]{Model}}} & \multicolumn{2}{c}{{\bf Test \BLEU}}  \\ %\cline{2-3} 
                         & {\bf IWSLT15 }     & {\bf WMT14 }    \\ %\cline{1-1}
\hline\hline
%\multicolumn{3}{l}{} \\
P1: $\mathcal{C}_{\text{random}}$            &      34.6    &     31.6               \\
P9: Curriculum (Dynamic) & {\bf 37.9} &   {\bf 35.6}  \\
P10: Static selection & 36.8 & 34.6 \\
\hline
%\multicolumn{3}{l}{} \\
W1: $\mathcal{C}_{\text{random}}$            &    36.5      &     35.0               \\
W9: Curriculum (Dynamic)  & {\bf 38.1 } & 36.3 \\
W10: Static selection & 37.4 & 36.2 \\
\end{tabular}
\end{center}
\caption{\small
Curriculum learning works slightly better than fine-tuning a warmed-up model with a top static selection.
\label{dynamic-vs-static}
}
\end{table}

What if we retrain on the static data, too? In Table~\ref{BLEU},
W11 vs. W9 shows that retrained models on the static data is far behind for the WMT14 training -- top 10\%
selection has only 4 million examples. On Paracrawl,  P11 vs. P9 are closer, but 
retraining on co-curriculum performs still better.
In all cases, co-curricular learning gives the best results.
We may tune the static selection for better results, but then it is
 the exact point of CL, to  evolve the data re-weighting without the need of a hard cutoff on
 selection ratio.

\begin{table}[t]
\small
\begin{center}
\begin{tabular}{l|cc}
%\hline
\multirow{2}{*}{{\bf \makecell[l]{Model}}} & \multicolumn{2}{c}{{\bf Test \BLEU}}  \\ %\cline{2-3} 
                         & {\bf IWSLT15 }     & {\bf WMT14 }    \\ %\cline{1-1}
\hline\hline
%\multicolumn{3}{l}{} \\
P9: Retrain with curriclum & {\bf 37.9} &   {\bf 35.6}  \\
P11: Retrain with static sel. & 37.1 & 34.6 \\
\hline
%\multicolumn{3}{l}{} \\
W9: Retrain with curriculum  & {\bf 38.1 } & {\bf 36.3} \\
%W12: ID fine-tune W9 & 39.9  & 36.9 \\
W11: Retrain on static sel. & 34.0 & 31.7 \\
\end{tabular}
\end{center}
\caption{\small
Curriculum learning works better than retraining with a static, top selection, especially when
the training dataset is small.
\label{BLEU}
}
\end{table}

\subsection{Discussion}

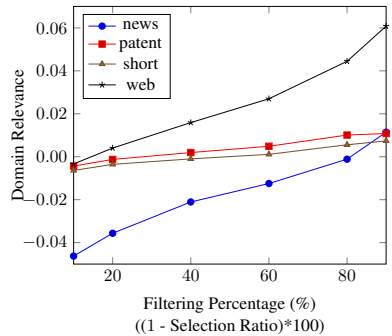
\begin{figure}[t]
\begin{center}
\begin{tikzpicture}[scale=0.6]
\begin{axis}[xlabel={\makecell[c]{Filtering Percentage (\%) \\ (($1$ - Selection Ratio)*100)}}, ylabel={Domain Relevance},
             ymax=0.07, ymin=-0.05, minor y tick num = 0,
             xmax=90,xmin=10,minor x tick num = 0,
             scaled y ticks = false,
             %cycle list name=black white,
             yticklabel style={%
        /pgf/number format/.cd,
            fixed,
            fixed zerofill,
            precision=2,
            /tikz/.cd
    },
                   legend pos=north west],
             ]
    \addplot table[x index=0,y index=1,col sep=comma] {quality_transfer.dat};
    \addlegendentry{news}
    \addplot table[x index=0,y index=2,col sep=comma] {quality_transfer.dat};
    \addlegendentry{patent}
    \addplot+[ mark=triangle*] table[x index=0,y index=3,col sep=comma] {quality_transfer.dat};
    \addlegendentry{short}
    \addplot table[x index=0,y index=4,col sep=comma] {quality_transfer.dat};
    \addlegendentry{web}
    
%\legend{Paracrawl,WMT}
\end{axis}
\end{tikzpicture}
\end{center}
\caption{\small Curriculum learning in one domain may enable curriculum learning in another. }
\label{TL}
\end{figure}

{\bf Evidence of data-quality transfer}. Figure~\ref{TL} visualizes that  CL in one domain (e.g., web)
may enable CL in another.  This is the foundation of our proposed method.
To draw the figure, using  a random sample of 2000  pairs from  WMT training data and some additional in-domain parallel data,  we sort examples by tightening the 
selection ratio according to a  true web curriculum. The web curve shows the co-relation between selection ratio and data relevance to web. The same data order appears to
yield increasing relevance to other domains, too, with bigger effect on a closer `news' domain, but smaller effect on `patent' and `short' (sentences).
% --
%Making the domain  monolingual corpus and the out-of-domain
%trusted parallel data closer in-domain may strengthen the effect of  co-curricular 
%learning.

%After being scored with the four $\phi$'s, each
%example  has
%four relevance scores.  We create a curriculum by sorting examples on the web scores,
%getting $\Phi(Q_t)$ (Eq.~\ref{ds_as_curri}), where $t$ is demonstrated
%using increase of percentile. In Figure~\ref{TL}, for web,  $\Phi(Q_t)$ increases
%as selection percentile gets tightened.  So does the wikinews curve, even
%though on the out-of-domain/web, relevance measure.

\vspace{0.1in}

\begin{comment}
\begin{figure}[t]
\begin{center}
\includegraphics[width=6cm,scale=0.6]{TL_curated.eps}
\end{center}
\caption{\small Curriculum learning in one domain may enable curriculum learning in another domain. \label{TL}}
\end{figure}
\end{comment}

\begin{comment}
\noindent
{\bf Choice of corpus.}  The curriculum learning effect is less obvious
for patent and short, which are further apart from web than wikinews.
So, in co-curricular learning, making the query corpus and the out-of-domain
trusted corpus closer in-domain would potentially enhance the co-curricular 
learning effect. In this sense, domain and denoising may be observations
at different extremes of a same quality measure.
\end{comment}

\noindent
{\bf Regularizing data without a teacher}. The analysis in Section~\ref{EM_effect} shows that 
the denoising scoring function and its bootstrapped versions tend to regularize
the late curriculum and make the scheduled data easier for small models to learn on.
One potential 
further application of this data property may be in learning a multitask
curriculum where regular data may be helpful for multiple task distributions
to work together in the same model. This has been achieved by knowledge distillation
in existing research \cite{tan2018multilingual}, by regularizing data with a teacher -- 
We could instead regularize data by example selection, without a teacher. We leave
this examination for future research.

\vspace{0.1in}

\noindent
{\bf Pace function hyper-parameters}. In experiments, we found that data-discarding pace functions
seem to work best when they simultaneously decay down to their respective floors.
%The cascaded co-curriculum uses hyper-parameters
%to control data-discarding paces, each for a different data-quality curriculum. In experiments,
Adaptively adjusting them  seems  an interesting future work.

\begin{comment}
\vspace{0.1in}

\noindent
{\bf ``Cleanness'' of data} We have been abusing the term ``cleanness'' to refer to
well-translated data. Very easy sentences can be ``clean'', but relative to
a strong system that can translate them better, they can become noisy.
bla bla bla.
\end{comment}

%\noindent
%{\bf Hyper-parameters} fsda dsf dfsd ff f f fdf sdf 

%\vspace{-0.05in}
\section{Conclusion}
We present a co-curricular learning method to make domain-data selection  work
better on  noisy data, by dynamically composing it with  clean-data selection.
We show that the method improves over either constituent selection and their static combination.
We further refine the co-curriculum with an EM-style optimization procedure and show its effectiveness, in particular on small-capacity models.
In future, we would like to extend the method to handle more than two curricula objectives.

 %Our work is built upon, and extends the single (final) task curriculum learning research in NMT.
 %Besides transfer learning of data quality, the co-curricular learning idea can potentially be
 %used to compose NMT curricula of other purposes. For example, when it tries to 
 %compose an easy-to-hard curriculum for faster convergence and a quality curriculum for accuracy,
 %the resulting co-curricula may achieve both in one training run. We wish our work could inspire more
 %research into widening curriculum learning's impact on NNT.

\section*{Acknowledgments}
The authors would like to thank Yuan Cao for his help and advice, the three anonymous reviewers for their constructive reviews, Melvin Johnson for early discussions,  Jason Smith, Orhan Firat, Macduff Hughes for comments on an earlier draft,
and Wolfgang Macherey for his early work on the topic.

%\noindent \textbf{Preparing References:} \\
%Include your own bib file like this:
%\verb|\bibliographystyle{acl_natbib}|
%\verb|\bibliography{acl2019}| 

%where \verb|acl2019| corresponds to a acl2019.bib file.
\bibliography{acl2019}
\bibliographystyle{acl_natbib}

\end{document}